\documentclass[runningheads]{llncs}

\usepackage{algorithm}

\usepackage{comment}
\usepackage{cite}
\usepackage{amsmath,amssymb,amsfonts}
\usepackage{graphicx}
\usepackage{textcomp}
\usepackage{xcolor}

\usepackage{algorithmicx}
\usepackage{algpseudocode}
\usepackage{graphicx}
\usepackage{subfigure}

\usepackage{enumitem}
\usepackage{booktabs}
\usepackage{multirow}

\def\BibTeX{{\rm B\kern-.05em{\sc i\kern-.025em b}\kern-.08em
    T\kern-.1667em\lower.7ex\hbox{E}\kern-.125emX}}

\graphicspath{{../figs/}}
\DeclareGraphicsExtensions{.pdf,.jpeg,.png}

\newcommand{\eat}[1]{}
\renewcommand{\paragraph}[1]{\smallskip\noindent {\bf #1}}
\newcommand{\indep}{\rotatebox[origin=c]{90}{$\models$}}

\begin{document}
\title{An Influence-based Approach for Root Cause Alarm Discovery in Telecom Networks}
\titlerunning{Root Cause Alarm Discovery in Telecom Networks}

\author{
Keli Zhang*, Marcus Kalander*, Min Zhou, Xi Zhang and Junjian Ye
}

\authorrunning{Zhang et al.}

\institute{Noah's Ark Lab, Huawei Technologies}

\maketitle

\begin{abstract}
Alarm root cause analysis is a significant component in the day-to-day telecommunication network maintenance, and it is critical for efficient and accurate fault localization and failure recovery. 
In practice, accurate and self-adjustable alarm root cause analysis is a great challenge due to network complexity and vast amounts of alarms.
A popular approach for failure root cause identification is to construct a graph with approximate edges, commonly based on either event co-occurrences or conditional independence tests. However, considerable expert knowledge is typically required for edge pruning.
We propose a novel data-driven framework for root cause alarm localization, combining both causal inference and network embedding techniques. In this framework, we design a hybrid causal graph learning method (HPCI), which combines Hawkes Process with Conditional Independence tests, as well as propose a novel Causal Propagation-Based Embedding algorithm (CPBE) to infer edge weights.
We subsequently discover root cause alarms in a real-time data stream by applying an influence maximization algorithm on the weighted graph.
We evaluate our method on artificial data and real-world telecom data, showing a significant improvement over the best baselines.

\keywords{Network Management \and Root Cause Analysis \and Alarm Correlation Analysis \and Influence Maximization}.
\end{abstract}
\def\thefootnote{*}\footnotetext{These authors contributed equally to this work.}
\vspace*{-0.5cm}
\section{Introduction}
\label{sec:introduction}
Recent years have seen rapid development in cellular networks,
both in increasing network scale and complexity coupled with increasing network performance demands.
This growth has made the quality of network management an even greater challenge and puts limits on the analysis methods that can be applied.
In cellular networks, anomalies are commonly identified through alarms. A large-scale network can generate millions of alarms during a single day. Due to the interrelated network structure, a single fault can trigger a flood of alarms from multiple devices. Traditionally, to recover after a failure, an operator will analyze all relevant alarms and network information. This can be a slow and time-consuming process. However, not all alarms are relevant. There exists a subset of alarms that are the most significant for fault localization. We denote these as root cause alarms, and our main goal is to intelligently identify these alarms.

There exist abundant prior research in the areas of Root Cause Analysis (RCA) and fault localization.
However, most proposed methods are highly specialized and take advantage of specific properties of the deployed network, either by using integrated domain knowledge or through particular design decisions~\cite{Bahl2007, causeInfer2014, ge2010grca}. A more general approach is to infer everything from the data itself. 

In our proposed alarm RCA system, we create an influence graph to model alarm relations. Causal inference is used to infer an initial causal graph, and then we apply a novel Causal Propagation-Based Embedding (CPBE) algorithm to supplement the graph with meaningful edge weights. 
To identify the root cause alarms, we build upon ideas in how influence propagates in social networks and view the problem as an influence maximization problem~\cite{kempe2003maximizing}, i.e., we want to discover the alarms with the largest influence. When a failure transpires, our system can automatically perform RCA based on the sub-graph containing the involved alarms and output the top-$K$ most probable root cause alarms.

In summary, our main contributions are as follows:
\begin{itemize}
  \item We design a novel unsupervised approach for root cause alarm localization that integrates casual inference and influence maximization analysis, making the framework robust to causal analysis uncertainty without requiring labels.
  \item We propose HPCI, a Hawkes Process-based Conditional Independence test procedure for causal inference.
  \item We further propose CPBE, a Causal Propagation-Based Embedding algorithm based on network embedding techniques and vector similarity to infer edge weights in causality graphs.
  \item Extensive experiments on a synthetic and a real-world citywide dataset show the advantages and usefulness of our proposed methods.
\end{itemize}
\vspace*{-0.2cm}
\section{Related work}
\vspace*{-0.15cm}
\label{sec:relatedwork}

\paragraph{Root cause alarms.}
There are various ways to discover alarm correlations and root cause alarms. Rules and experience of previous incidents are frequently used. In more data-driven approaches, pattern mining techniques that compress alarm data can assist in locating and diagnosing faults~\cite{zhang2018network}.
Abele et al.~\cite{abele2013combining} propose to find root cause alarms by combining knowledge modeling and Bayesian networks.
To use an alarm clustering algorithm that considers the network topology and then mine association rules to find root cause alarms was proposed in~\cite{su2017association}.

\paragraph{Graph-based root cause analysis.}
Some previous works depend on system dependency graphs, e.g., Sherlock~\cite{Bahl2007}. A disadvantage is the requirement of exact conditional probabilities, which is impractical to obtain in large networks. Other systems are based on causality graphs. G-RCA~\cite{ge2010grca} is a diagnosis system, but its causality graph is configured by hand, which is unfeasible in large scale, dynamic environments. The PC algorithm~\cite{pc-algo-1991} is used by both CauseInfer~\cite{causeInfer2014} and \cite{mininglog2018} to estimate DAGs, which are then used to infer root causes. However, such graphs can be very unreliable. Co-occurrence and Bayesian decision theory are used in~\cite{lou2010mining} to estimate causal relations, but it is mainly based on log event heuristics and is hard to generalize. Nie et al.~\cite{CausalityGraph} use FP-Growth and lag correlation to build a causality graph with edge weights added with expert feedback.
\section{Preliminaries}
\label{sec:preliminaries}
In this section, we shortly review the two key concepts that our proposed method depends upon, Hawkes process~\cite{hawkes1971spectra} and the PC algorithm~\cite{spirtes2000causation}.

\paragraph{Hawkes Process.} This is a popular method to model continuous-time event sequences where past events can excite the probability of future events. The keystone of Hawkes process is the conditional intensity function, which indicates the occurrence rate of future events conditioned on past events, denoted by $\lambda_d(t)$, where $u \in \mathcal{C}=\{1,2,...,U\}$ is an event type. Formally, given an infinitely small time window $[t,t+\Delta t)$, the probability of a type-$u$ event occurring in this window is $\lambda_d(t)\Delta t$. For $U$-dimensional Hawkes process with event type set $\mathcal{C}$, each dimension $u$ has a specific form of conditional intensity function defined as
\begin{equation}
\small
\lambda_u(t) = \mu_u +\sum_{v \in \mathcal{C}}\sum_{t_i < t}k_{uv}(t-t_i),
\end{equation}
where $\mu_u \ge 0$ is the background intensity for type-$u$ events and $k_{uv}(t) \ge 0$ is a kernel function indicating the influence from past events. An exponential kernel is most frequently used, i.e., $k_{uv}(t) = \alpha_{uv}e^{-\beta_{uv}(t)}$, where $\alpha_{uv}$ captures the degree of influence of type-$v$ events to type-$u$ events and $\beta_{uv}$ controls the decay rate. The parameters are commonly learned by optimizing a log-likelihood function. Let $\mu = (\mu_u) \in R^U$ be the background intensities, and $A=(\alpha_{uv}) \in R^{U \times U}$ the influence matrix reflecting the certain causality between event types. For a set of event sequences $\mathcal{S}=\{S_1,S_2,...,S_m\}$, where each event sequence $S_i=\{(a_{ij},t_{ij})\}_{j=1}^{n_i}$ is observed during a time period of $[0,T_i]$, and each pair $((a_{ij},t_{ij}))$ represents an event of type $a_{ij}$ that occurred at time $t_{ij}$. The log-likelihood of a Hawkes process model with parameters $\Theta=\{\mu,A\}$ can then be expressed as
\begin{equation}
\small
\mathcal{L}(\mu,A) = \sum_{i=1}^m(\sum_{j=1}^{n_i}\log\lambda_{a_{ij}}(t_{ij}) - \sum_{u=1}^U\int_0^{T_i}\lambda_u(t)dt).
\end{equation}
The influence matrix $A$ is generally sparse or low-rank in practice, hence, adding penalties into $\mathcal{L}(\mu, A)$ is common. For instance, Zhou~\cite{zhou2013learning} used a mix of Lasso and nuclear norms to constrain $A$ to be both low-rank and sparse by using
\begin{equation}
\label{eq:hawkes}
\small
\min_{\mu \ge 0, A \ge 0}-\mathcal{L}(\mu,A) + \rho_1||A||_1 +\rho_2||A||_*,
\end{equation}
where $||\cdot||_1$ is the $L_1$-norm, and $||\cdot||_*=\sum_{i=1}^{rankA}\sigma_i$ is the nuclear norm. The parameters $\rho_1$ and $\rho_2$ controls their weights. A number of algorithms can be applied to solve the above learning problem, more details can be found in~\cite{veen2008estimation}.

\paragraph{PC Algorithm.} This algorithm is frequently used for learning directed acyclic graphs (DAGs) due to its strong causal structure discovery ability~\cite{wang2018cloudranger}. Conditional Independence (CI) tests play a central role in the inference. A significance level $p$ is used as a threshold to determine if an edge should be removed or retained. Formally, given a variable set $Z$, if $X$ is independent of $Y$, denoted as $X \indep Y | Z$, the edge between $X$ and $Y$ will be removed, otherwise it will be kept in the causal graph. A rigorous description can be found in~\cite{kalisch2007estimating}.

The G-square test and Fisher-Z test are two common realizations for conditional independence testing in causal inference~\cite{peters2014causal}. The G-square test is used for testing independence of categorical variables using conditional cross-entropy while the Fisher-Z test evaluates conditional independence based on Pearson's correlation. CI tests assume that the input is independent and identically distributed. In our alarm RCA scenario, the size of the time window depends on the network characteristics and needs to be selected to ensure that causal alarms exist in one window and the data between different windows are independent.

\section{System Overview}
\label{sec:overview}
\begin{figure}[t!]
    \centering
    \includegraphics[width=0.75\textwidth, trim={5pt 0 20pt 0}, clip]{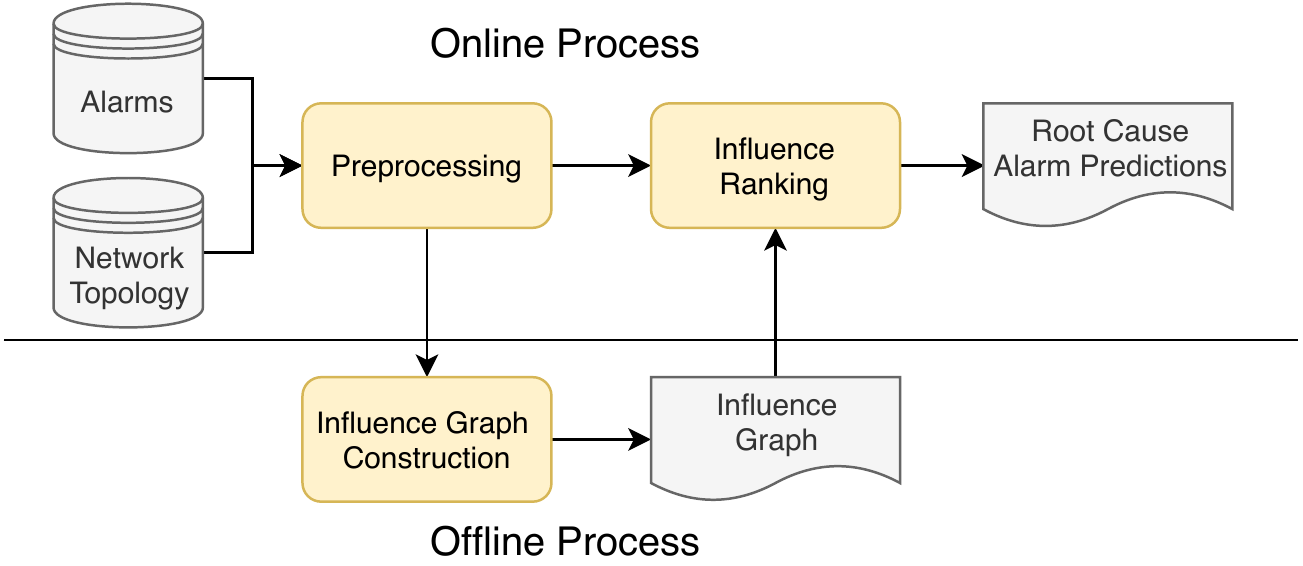}
    \vspace{-0.3cm}
    \caption{Architecture of the proposed system.}
    \vspace{-0.5cm}
    \label{fig:framework}
\end{figure}

Our proposed framework consists of two main procedures: influence graph creation and alarm ranking. A system overview can be found in Figure~\ref{fig:framework}.
The alarm preprocessing module is shared and handles alarm filtering and aggregation with consideration to the network topology. 

The influence graph is constructed using historical alarm transactions and is periodically recreated. It is comprised of alarm types as nodes and their inferred relations as edges.
To create the graph, we first exploit causal inference methodology to infer an initial alarm causality graph structure by applying HPCI, a hybrid method that merges Hawkes process and conditional independence tests. We further apply a network embedding technique, CPBE, to infer the edge weights. The alarm stream is monitored in real-time. When a failure transpires, the system attempts to discover the underlying root cause alarms. The related alarms are aggregated with the created influence graph and are ranked by their influence to determine the top-$K$ most probable root cause alarm candidates. The alarm candidates are then given to the network operators to assist in handling the network issue.
\section{Methodology}
\label{sec:method}
This section introduces the key components in our system; alarm preprocessing, the influence graph construction, and how the influence ranking is done. We start by presenting the data and its required preprocessing and aggregation steps.

\subsection{Data and Alarm Preprocessing}
\label{subsec:preprocess}
\paragraph{Network Topology.}
This is the topological structure of the connections between network devices. Connected network devices will interact with each other. If a failure occurs on one device, then any connected devices can be affected, triggering alarms on multiple network nodes.

\paragraph{Alarms.} 
Network alarms are used to identify faults in the network. Each alarm record contains information about occurrence time, network device where the alarm originated, alarm type, etc. In practice, any alarms with missing key information are useless and removed. Furthermore, alarm types that are either systematic or highly periodical are also removed. These types of alarms are irrelevant for root cause analysis since they will be triggered regardless if a fault occurred or not.

\paragraph{Alarm Preprocessing.}
We partition the raw alarm data into alarm sequences and alarm transactions in three steps as follows.
\begin{enumerate}
    \item Devices in connected sub-graphs of the network can interact, i.e., alarms from these devices can potentially be related to the same fault. Consequently, we first aggregate alarms from the same sub-graph together.
    \item Alarms related to the same failure will generally occur together within a short time interval. 
    We thus further partition the alarms based on their occurrence times. Alarms that occurred within the same time window $w_i$ are grouped and sorted by time. The window size can be adjusted depending on network characteristics. 
    We define each group as an alarm sequence, denoted as $S_i=\{(a_{ij},t_{ij})\}_{j=1}^{n_i}$, where $w_i$ is the window, $a_{ij} \in A$ is the alarm type, $t_{ij} \in w_i$ is occurrence time, and $n_i$ the number of alarms.
    \item Each alarm sequence is transformed into an alarm transaction denoted by $T_i =\{(a,t,n)|a \in A_i\}$, where $a,t,n$ indicates the alarm type, the earliest occurrence time and the number of occurrences, respectively. Different from $S_i$, $A_i$ contains a single element for each alarm type in window $w_i$.
\end{enumerate}

\subsection{Alarm Influence Graph Construction}
\label{sec:graph-construct}
In this section, we elaborate on the construction of the alarm influence graph. The graph has the alarm types as nodes and their relation as the edges. First, an initial causal structure DAG is inferred by a hybrid causal structure learning method (HPCI). Subsequently, edge weights are inferred using a novel network embedding method (CPBE).

\paragraph{HPCI.}
A multi-dimensional Hawkes process can capture certain causalities behind event types, i.e., the transpose of the influence matrix $A$ can be seen as the adjacency matrix of the causal graph for event types. However, redundant or indirect edges tend to be discovered since the conditional intensity function can not perfectly model real-world data and due to the difficulty in capturing the instantaneous causality.

To reduce this weakness, we propose a hybrid algorithm HPCI that is based on Hawkes process and the PC algorithm. HPCI is used to discover the causal structure for the alarm types in our alarm RCA scenario.
The main procedure can be expressed in three steps.
(1) Use multi-dimensional Hawkes process without penalty to capture the influence intensities among the alarm types. We use the alarm sequences $\mathcal{S}=\{S_i\}$ as input and obtain an initial weighted graph. The weights on an edge $(u,v)$ is the influence intensity $\alpha_{uv} > 0$, reflecting the expectation of how long it takes for a type-$u$ event to occur after an type-$v$ event. All edges with positive weights are retained.
(2) Any redundant and indirect causal edges are removed using CI tests. We use the alarm transactions $\mathcal{T}=\{T_i\}$ as input and for each alarm $a_i$ the sequence of alarm occurrences $N_i = \{n| T_k \in \mathcal{T}, (a,t,n) \in T_k, a = a_i\}$ is extracted. Note that $n$ can be $0$ if an alarm type is not present in a window $w_i$. For each pair of alarm types $(a_i, a_j)$, the CI test of their respective occurrence sequences is used to test for independence and remove edges. The output is a graph with unwanted edges removed. 
(3) Finally, we iteratively remove the edge with the smallest intensity until the graph is a DAG. Our final causal graph is denoted as $G^C$. 

We select CI tests to enforce sparsity in the causal graph in the second step. Compared to adding penalty terms such as $L1$-norm, the learning procedure is more interpretable, and our experiments show more robust results. 

\paragraph{Edge Weights Inference.}
The causal graph $G^C$ learned by HPCI is a weighted graph, however, the weights do not account for global effects on the causal intensities. Hence, to encode more knowledge into the graph, we propose a novel network embedding-based weight inference method, Causal Propagation-Based Embedding (CPBE). CPBE consists mainly of two steps; (1) For each node $u$, we obtain a vector representation $Z_u \in \mathcal{R}^L$ using a novel network embedding technique. (2) Use vector similarity to compute edge weights between nodes.

\begin{figure}[t!]
	\centering
	\includegraphics[width=1.0\linewidth]{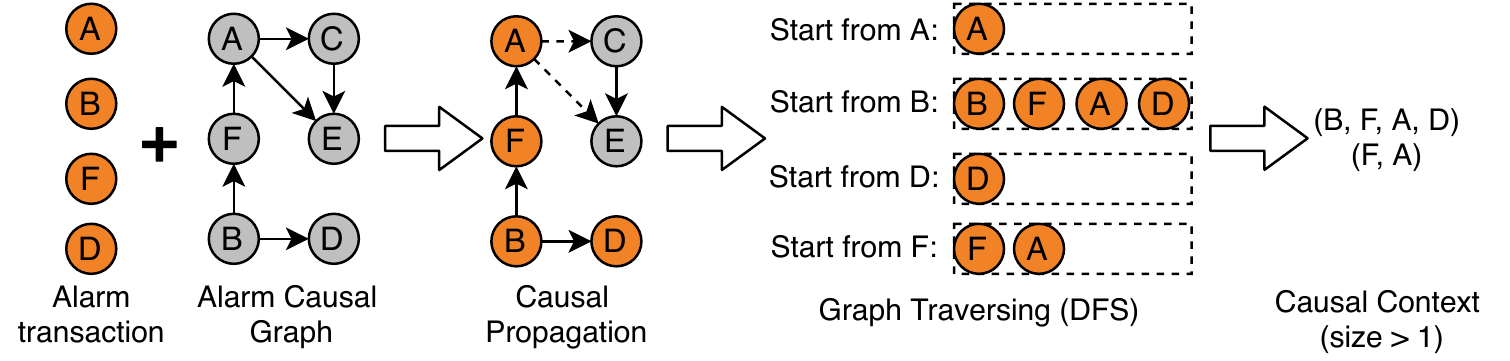}
	\vspace{-0.4cm}
	\caption{Context generation procedure for CPBE.}
	\label{fig:causal-context-gen}
	\vspace{-0.4cm}
\end{figure}

The full CPBE algorithm is shown in Algorithm~\ref{alg:cpbe}. CPBE uses a new procedure to generate a context for the skip-gram model~\cite{mikolov2013distributed} (lines 1-9). This procedure is also illustrated in Figure~\ref{fig:causal-context-gen}.
In essence, for each historical alarm transaction $T_i \in \mathcal{T}$, we use the learned causality graph $G^C$ and extract a causal propagation graph $G^{PC}_i$, where only the nodes corresponding to alarm types in $T_{i}$ are retained. Starting from each node in $G^{PC}_i$, we traverse the graph to generate a node-specific causal context. During the traversal for a node $u$, only nodes that have a causal relation with $u$ are considered. There are various possible traversing strategies, e.g., depth-first search (DFS) and RandomWalk~\cite{node2vec}. The skip-gram model is applied to the generated contexts to obtain an embedding vector $Z_u \in \mathcal{R}^L$ for each node $u$. Finally, the edge weight between two nodes is set to be the cosine similarity of their associated vectors. We denote the final weighted graph as the alarm influence graph $G^I$.

\begin{algorithm}[t]
    \small
	\caption{Causal Propagation-Based Embedding (CPBE)}
	\label{alg:cpbe}
	\textbf{Input:} Alarm Transactions $\mathcal{T}=\{T_i\}$; Causal Graph $G^C=(V,E)$;
	\hspace*{\algorithmicindent} 
	\begin{algorithmic}[1]
	\State $C = \{\}, E_{w} = \{\}$;
    \For{ $T_i\in T$}:
    \State $G^{PC}_i = $ ConstructPropagationGraph$(T_{i},G^C)$
    \For{$alarm\_node \in T_{i}$}:
    \State $C_{alarm\_node} = $ GraphTraversing$(G^{PC}_i,alarm\_node)$
    \State $C = C \cup C_{alarm\_node}$
    \EndFor
	\EndFor
	\State $Z \gets \text{ skip-gram($C$) to map nodes to embedding vectors}$
	\For{ $(u,v) \in E$}:
	\State $w = Cosine(Z_u, Z_v)$
	\State $E_{w} = E_{w} \cup (u,v,w)$
	\EndFor
	\end{algorithmic}
	\textbf{Output:} Alarm Influence Graph $G^I=(V,E_{w})$;
\end{algorithm}

\subsection{Root Cause Alarm Influence Ranking}
\label{sec:influence-rank}
\begin{figure}[t!]
	\centering
	\includegraphics[width=1.0\linewidth]{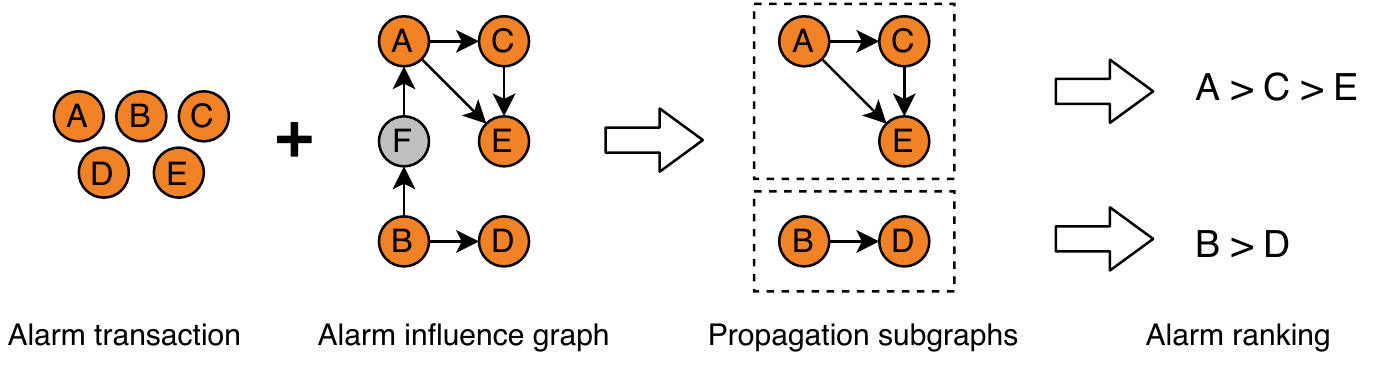}
	\vspace{-0.6cm}
	\caption{Processing flow from alarm transaction to ranked alarms.}
	\label{fig:influence-rank-online}
	\vspace{-0.5cm}
\end{figure}

This section describes how the alarm influence graph $G^I$ is applied to an alarm transaction to identify the root cause alarms. For each alarm transaction $T_i \in \mathcal{T}$, an alarm propagation graph $G^{PI}_i$ is created with the relevant nodes $v \in T_i$ and applicable edges $\{(u,v,w) | u,v \in T_i\}$. Any nodes corresponding to alarms not present in $T_i$ are removed. The process is equivalent to how $G^{PC}_i$ is created from the causal graph $G^C$. The alarms in each propagation sub-graph are then ranked independently. The process is illustrated in Figure~\ref{fig:influence-rank-online}. 

We consider the problem of finding the root cause alarm as an influence maximization problem~\cite{kempe2003maximizing}. We want to discover a small set of $K$ \textit{seed nodes} that maximizes the influence spread under an influence diffusion model. A suitable model is the independent cascade model, which is widely used in social network analysis.
Following this model, each node $v$ is activated by each of its neighbors independently based on an influence probability $p_{u,v}$ on each edge $(u,v)$. These probabilities directly correspond to the learned edge weights. 
Given a seed set $S_0$ to start with at $t=0$, at step $t > 0$, $u \in S_{t-1}$ tries to activate its outgoing inactivated neighbors $v \in \mathcal{N}^{out}(u)$ with probability $p_{u,v}$. Activated nodes are added to $S_t$ and the process terminates when $|S_t| = 0$, i.e., when no nodes further nodes are activated. The influence of the seed set $S_0$ is then the expected number of activated nodes when applying the above stochastic activation procedure.

There are numerous algorithms available to solve the influence maximization problem~\cite{li2018influence}. In our scenario, each graph $G^{PI}_i$ is relatively small and the actual algorithm is thus less important. We directly select the Influence Ranking Influence Estimation algorithm (IRIE)~\cite{Jung2012} for this task.
IRIE estimates the influence $r(u)$ for each node $u$ by deriving a system of $n$ linear equations with $n$ variables. The influence of a node $u$ comprises of its own influence, $1$, and the sum of the influences it propagates to its neighbors.

\section{Evaluation}
\label{sec:evaluation}
In this section, we present the experimental setup and evaluation results. We perform two main experiments, one to verify the correctness of our causal graph and a second experiment to evaluate the root cause identification accuracy. The first experiment is performed on both synthetic and real-world data, while the second is completed on the real-world dataset. The datasets and code are available at https://github.com/shaido987/alarm-rca.

\paragraph{Synthetic Data Generation.}
The synthetic event sequences are generated in four steps. 
(1) We randomly generate a DAG $G$ with an average out-degree $d$ with $N$ event types. We set $d$ to $1.5$ to emulate the sparsity property of our real-world dataset.
(2) For each edge $(u,v)$, a weight $\alpha_{uv}$ is assigned by uniform random sampling from a range $r \in [(0.01, 0.05),(0.05, 0.1), (0.1,0.5),(0.5,1.0)]$.
(3) For each event type $u \in U$, we assign a background intensity $\mu_u$ by uniform random sampling from $(0.001, 0.005)$.
(4) Following Ogata~\cite{ogata1981lewis}, we use $\alpha_{uv}$ and $\mu_u$ as parameters of a Multi-dimensional Hawkes process and simulate event sequences. We generate event sequences of length $T=14$ days while ensuring that the total number of events is greater than $10{,}000$.

\paragraph{Real-world Dataset.} 
The dataset was collected from a major cellular carrier in a moderate-sized city in China between Aug 4th, 2018 and Oct 24th, 2018. After preprocessing, it consists of $672{,}639$ alarm records from $3{,}818$ devices with $78$ different alarm types. Due to the difficulty of labeling causal relations, we only have the ground-truth causal relations for a subset of $15$ alarm types, $44$ directed edges in the graph. Furthermore, we have also obtained the ground-truth root cause alarms in a random sample of $6{,}000$ alarm transactions. These are used to evaluate the root cause localization accuracy.

\subsection{Causal Graph Structure Correctness}
We evaluate our proposed HPCI method and the accuracy of the discovered causal graphs. We use four frequently used causal inference methods for sequential data as baselines.
\begin{itemize}
\item PC-GS: PC algorithm with G-square CI test.
\item PC-FZ: PC algorithm with Fisher-Z CI test.
\item PCTS: Improved PC algorithm for causal discovery in time series~\cite{meng2020localizing}.
\item HPADM4: Multi-dimensional Hawkes process with exponential parameterization of the kernels and a mix of $L1$ and nuclear-norm~\cite{zhou2013learning}.
\end{itemize}
The significance level $p$ in the conditional independence tests included in the methods are all set to $0.05$. The size of time window $w$ for aggregating event sequences is set to $300$ seconds, the maximum lag $\tau_{max}=2$ in PCTS, and the penalization level in HPADM4 is set to the default $1{,}000$.
Furthermore, the decay parameter $\beta$ in Hawkes process is set to $0.1$, and we select Fisher-Z as the CI test in our HPCI algorithm.
For evaluation, we define three metrics as follows.
\begin{equation*}
\small
  Precision = \frac{|P \cap S|}{|P|}, \quad
  Recall = \frac{|P \cap S|}{|S|}, \quad
  F1{\text -}score = 2 \cdot \frac{Precision \cdot Recall}{Precision + Recall},
\end{equation*}
where $P$ is the set of all directed edges in the learned causal graph $G^C$ and $S$ is the set of ground-truth edges. 

\paragraph{Results.}
\begin{table}[t]
\setlength\tabcolsep{11pt} 
\centering
\caption{F1-scores on the synthetic dataset for different number of event types and $\alpha$.}
\label{tab:synthetic-results}
\begin{tabular}{ccccccc}
\toprule
$\alpha$ & N & PS-GS & PS-FZ & PCTS & HPADM4 & HPCI \\
\midrule
\multirow{3}{*}{(0.01,0.05)}  & 10  & 0.286 & 0.174 & 0.283 & $\mathbf{0.566}$ & 0.200 \\
                              & 20  & 0.133 & 0.321 & 0.156 & 0.306 & $\mathbf{0.604}$ \\
                              & 30  & 0.216 & 0.267 & 0.104 & 0.229 & $\mathbf{0.357}$ \\
                    \midrule
\multirow{3}{*}{(0.05,0.1)}  & 10  & 0.500 & 0.529 & 0.283 & 0.500 & $\mathbf{0.867}$ \\
                             & 20  & 0.367 & 0.585 & 0.155 & 0.323 & $\mathbf{0.806}$ \\
                             & 30  & 0.265 & 0.484 & 0.227 & 0.227 & $\mathbf{0.756}$ \\
                    \midrule
\multirow{3}{*}{(0.1,0.5)}  & 10  & 0.467 & 0.811 & 0.278 & 0.517 & $\mathbf{0.933}$  \\
                            & 20  & 0.621 & 0.889 & 0.151 & 0.306 & $\mathbf{0.984}$  \\
                            & 30  & 0.495 & 0.845 & 0.103 & 0.227 & $\mathbf{0.967}$  \\
                    \midrule
\multirow{3}{*}{(0.5,1.0)}  & 10  & 0.800 & 0.722 & 0.272 & 0.517 & $\mathbf{0.929}$ \\
                            & 20  & 0.708 & 0.906 & 0.151 & 0.302 & $\mathbf{0.983}$ \\
                            & 30  & 0.433 & 0.845 & 0.103 & 0.227 & $\mathbf{0.967}$ \\
                    \bottomrule
\end{tabular}
\end{table}
The F1-scores using synthetic data with $N \in [10,15,20]$ are shown in Table~\ref{tab:synthetic-results}. As shown, HPCI outperforms the baselines for nearly all settings of $N$ and $\alpha$. However, HPADM4 obtains the best result for $N=10$ and low $\alpha$, this is due to the distribution of event occurrence intervals being sparse which makes the causal dependency straightforward to capture using a Hawkes process. However, for higher $N$ or $\alpha$ the events will be denser. Thus, Hawkes process has trouble distinguishing instantaneous causal relations, especially when events co-occur. The use of CI tests in HPCI helps to distinguish these instantaneous causal relations by taking another perspective in which causality is discovered based on distribution changes in the aggregated data without considering the time-lagged information among events. HPCI thus achieves better results. The use of time aggregation is disadvantageous for PCTS due to its focus on time series, which can partly explain its comparatively worse results.

The results on the real-world data are shown in Table~\ref{tab:real-comparsion}. 
HPCI performs significantly better than all baselines in precision and F1-score, while PTCS obtains the highest recall.
PTCS also has significantly lower precision, indicating more false positives. PCTS is designed for time series, however, those may be periodic, which can give higher lagged-correlation values leading to more redundant edges.
HPCI instead finds a good balance between precision and recall. The competitive result indicates that the causality behind the real alarm data conforms to the assumptions of HPCI to a certain extent. 
\begin{table}[t]
\centering
\caption{Results of the causal graph structure evaluation on the real-world dataset.}
\label{tab:real-comparsion}
\begin{tabular}{lccc}
\toprule
Method & Precision      & Recall         & F1-score       \\ 
\midrule
PC-GS  & 0.250          & 0.159          & 0.194          \\
PC-FZ  & 0.452          & 0.432          & 0.442          \\
PCTS   & 0.220          & \textbf{0.864}          & 0.350          \\
HPADM4 & 0.491          & 0.614 & 0.545          \\
HPCI   & \textbf{0.634} & 0.591 & \textbf{0.612} \\ \bottomrule
\end{tabular}
\vspace{-0.3cm}
\end{table}
\subsection{Root Cause Alarm Identification}
We evaluate the effectiveness of CPBE and the root cause alarm accuracy on the real-world dataset. We use the causal graph structure created by HPCI as the base and augment it with the $44$ known causal ground-truths. The causal graph is thus as accurate as possible. CPBE is compared with four baseline methods, all used for determining edge weights.
\begin{itemize}
	\item IT, directly use the weighted causal graph discovered by HPCI with the learned influence intensities as edge weights.
    \item Pearson, uses the aligned Pearson correlation of each alarm pair~\cite{CausalityGraph}.
    \item CP, the weights of an edge $(u,v)$ is set to $\frac{A_{uv}}{A_u}$ where $A_{uv}$ is the number of times $u$ and $v$ co-occur in a window, and $A_u$ is the total number of $u$ alarms.
    \item ST, a static model with maximization likelihood estimator~\cite{goyal2010learning}. It is similar to CP, but $A_{uv}$ represents the number of times $u$ occurs before $v$.
\end{itemize}
For each method, IRIE is used to find the top-$K$ most likely root cause alarms in each of the $6{,}000$ labeled alarm transactions. For IRIE, we use the default parameters. We attempt to use RandomWalk, BFS, and DFS for traversal in CPBE, as well as different Skip-gram configurations with $w \in [1,5]$ and vector length $L \in [10, 30]$. However, there is no significant difference in the outcome, indicating that CPBE is insensitive to these parameter choices on our data. The results for different $K$ when using RandomWalk are shown in Table~\ref{tab:rca-comparsion}. As shown, CPBE outperforms the baselines for all $K$. For $K=1$, CPBE achieves an accuracy of 61.8\% which, considering that no expert knowledge is integrated into the system, is an excellent outcome. Moreover, the running time of CPBE is around $10$ seconds and IRIE takes $325$ seconds for all $6{,}000$ alarm transactions. This is clearly fast enough for system deployment.
\begin{table}[t]
\centering
\caption{Root cause alarm identification accuracy using different edge weight inference strategies together with IRIE for alarm ranking at different $K$.}
\label{tab:rca-comparsion}
\begin{tabular}{lccccc}
\toprule
Method  & K=1         & K=2         & K=3         & K=4                     & K=5         \\
\midrule
IT      & 0.576          & 0.590          & 0.672          & 0.810                       & 0.900           \\
Pearson & 0.407          & 0.435          & 0.456          & 0.486                      & 0.486          \\
CP      & 0.474          & 0.640          & 0.730          & 0.790                      & 0.840           \\
ST      & 0.439          & 0.642          & 0.750          & 0.785 & 0.814          \\
CPBE    & \textbf{0.618} & \textbf{0.752} & \textbf{0.851} & \textbf{0.929}             & \textbf{0.961} \\
\bottomrule
\end{tabular}
\vspace{-0.4cm}
\end{table}
\section{Conclusion}
\label{sec:conclusion}
We present a framework to identify root cause alarms of network faults in large telecom networks without relying on any expert knowledge. We output a clear ranking of the most crucial alarms to assist in locating network faults. 
To this end, we propose a causal inference method (HPCI) and a novel network embedding-based algorithm (CPBE) for inferring network weights. Combining the two methods, we construct an alarm influence graph from historical alarm data.
The learned graph is then applied to identify root cause alarms through a flexible ranking method based on influence maximization.
We verify the correctness of the learned graph using known causal relation and show a significant improvement over the best baseline on both synthetic and real-world data.
Moreover, we demonstrate that our proposed framework beat the baselines in identifying root cause alarms.

\bibliographystyle{splncs04}
\bibliography{paper}

\begin{thebibliography}{10}
\providecommand{\url}[1]{\texttt{#1}}
\providecommand{\urlprefix}{URL }
\providecommand{\doi}[1]{https://doi.org/#1}

\bibitem{abele2013combining}
Abele, L., Anic, M., et~al.: Combining knowledge modeling and machine learning
  for alarm root cause analysis. IFAC Proceedings Volumes  \textbf{46}(9),
  1843--1848 (2013)

\bibitem{Bahl2007}
Bahl, P., Chandra, R., et~al.: Towards highly reliable enterprise network
  services via inference of multi-level dependencies. In: ACM SIGCOMM Computer
  Communication Review. vol.~37, pp. 13--24. ACM (2007)

\bibitem{causeInfer2014}
Chen, P., Qi, Y., et~al.: Causeinfer: automatic and distributed performance
  diagnosis with hierarchical causality graph in large distributed systems. In:
  INFOCOM, 2014 Proceedings IEEE. pp. 1887--1895. IEEE (2014)

\bibitem{ge2010grca}
Ge, Z., Yates, J., et~al.: Grca: A generic root cause analysis platform for
  service quality management in large isp networks. In: ACM ACM Conference on
  Emerging Networking Experiments and Technologies (2010)

\bibitem{goyal2010learning}
Goyal, A., Bonchi, F., et~al.: Learning influence probabilities in social
  networks. In: Proceedings of the third ACM international conference on Web
  search and data mining. pp. 241--250. ACM (2010)

\bibitem{node2vec}
Grover, A., Leskovec, J.: node2vec: Scalable feature learning for networks. In:
  Proceedings of the 22nd ACM SIGKDD international conference on Knowledge
  discovery and data mining. pp. 855--864. ACM (2016)

\bibitem{hawkes1971spectra}
Hawkes, A.G.: Spectra of some self-exciting and mutually exciting point
  processes. Biometrika  \textbf{58}(1),  83--90 (1971)

\bibitem{Jung2012}
Jung, K., Heo, W., et~al.: Irie: Scalable and robust influence maximization in
  social networks. In: Data Mining (ICDM), 2012 IEEE 12th International
  Conference on. pp. 918--923. IEEE (2012)

\bibitem{kalisch2007estimating}
Kalisch, M., B{\"u}hlmann, P.: Estimating high-dimensional directed acyclic
  graphs with the {PC}-algorithm. Journal of Machine Learning Research
  \textbf{8},  613--636 (2007)

\bibitem{kempe2003maximizing}
Kempe, D., Kleinberg, J., et~al.: Maximizing the spread of influence through a
  social network. In: Proceedings of the ninth ACM SIGKDD international
  conference on Knowledge discovery and data mining. pp. 137--146. ACM (2003)

\bibitem{mininglog2018}
Kobayashi, S., Otomo, K., et~al.: Mining causality of network events in log
  data. IEEE Transactions on Network and Service Management  \textbf{15}(1),
  53--67 (2018)

\bibitem{li2018influence}
Li, Y., Fan, J., et~al.: Influence maximization on social graphs: A survey.
  IEEE Transactions on Knowledge and Data Engineering  (2018)

\bibitem{lou2010mining}
Lou, J.G., Fu, Q., et~al.: Mining dependency in distributed systems through
  unstructured logs analysis. SIGOPS Operating Systems Review  \textbf{44}(1),
  91--96 (2010)

\bibitem{meng2020localizing}
Meng, Y., Zhang, S., Sun, Y., Zhang, R., Hu, Z., Zhang, Y., Jia, C., Wang, Z.,
  Pei, D.: Localizing failure root causes in a microservice through causality
  inference. In: 2020 IEEE/ACM 28th International Symposium on Quality of
  Service (IWQoS). pp. 1--10. IEEE (2020)

\bibitem{mikolov2013distributed}
Mikolov, T., Sutskever, I., et~al.: Distributed representations of words and
  phrases and their compositionality. In: Advances in neural information
  processing systems. pp. 3111--3119 (2013)

\bibitem{CausalityGraph}
Nie, X., Zhao, Y., et~al.: Mining causality graph for automatic web-based
  service diagnosis. In: Performance Computing and Communications Conference
  (IPCCC), 2016 IEEE 35th International. pp.~1--8 (2016)

\bibitem{ogata1981lewis}
Ogata, Y.: On lewis' simulation method for point processes. IEEE transactions
  on information theory  \textbf{27}(1),  23--31 (1981)

\bibitem{peters2014causal}
Peters, J., Mooij, J.M., et~al.: Causal discovery with continuous additive
  noise models. The Journal of Machine Learning Research  \textbf{15}(1),
  2009--2053 (2014)

\bibitem{pc-algo-1991}
Spirtes, P., Glymour, C.: An algorithm for fast recovery of sparse causal
  graphs. Social science computer review  \textbf{9}(1),  62--72 (1991)

\bibitem{spirtes2000causation}
Spirtes, P., Glymour, C.N., et~al.: Causation, prediction, and search. MIT
  press (2000)

\bibitem{su2017association}
Su, C., Hailong, Z., et~al.: Association mining analysis of alarm root-causes
  in power system with topological constraints. In: Proceedings of the 2017
  International Conference on Information Technology. pp. 461--468. ACM (2017)

\bibitem{veen2008estimation}
Veen, A., Schoenberg, F.P.: Estimation of space--time branching process models
  in seismology using an em--type algorithm. Journal of the American
  Statistical Association  \textbf{103}(482),  614--624 (2008)

\bibitem{wang2018cloudranger}
Wang, P., Xu, J., et~al.: Cloudranger: Root cause identification for cloud
  native systems. In: 2018 18th IEEE/ACM International Symposium on Cluster,
  Cloud and Grid Computing (CCGRID). pp. 492--502. IEEE (2018)

\bibitem{zhang2018network}
Zhang, X., Bai, Y., et~al.: Network alarm flood pattern mining algorithm based
  on multi-dimensional association. In: Proceedings of the 21st ACM
  International Conference on Modeling, Analysis and Simulation of Wireless and
  Mobile Systems. pp. 71--78. ACM (2018)

\bibitem{zhou2013learning}
Zhou, K., Zha, H., et~al.: Learning social infectivity in sparse low-rank
  networks using multi-dimensional hawkes processes. In: Artificial
  Intelligence and Statistics. pp. 641--649 (2013)

\end{thebibliography}

\end{document}